\pdfoutput=1

\documentclass[11pt]{article}

\usepackage{emnlp2021}

\usepackage{times}
\usepackage{latexsym}

\usepackage{multirow}
\usepackage{graphicx}
\usepackage{subfigure}
\usepackage{tabu}
\usepackage{enumitem}
\usepackage{amsmath}
\usepackage{hyperref}
\DeclareMathOperator*{\argmax}{argmax} 
\usepackage{booktabs}
\usepackage{hyperref}

\usepackage[T1]{fontenc}

\usepackage[utf8]{inputenc}

\usepackage{microtype}

\title{Weakly-Supervised Visual-Retriever-Reader for Knowledge-based Question Answering}

\author{
Man Luo\footnotemark[1] \quad Yankai Zeng\footnotemark[1] \quad Pratyay Banerjee \quad Chitta Baral  \\
Arizona State University \\
\texttt{\{mluo26, yzeng55, pbanerj6, chitta\}@asu.edu}               \\
}

\begin{document}
\maketitle
\renewcommand{\thefootnote}{\fnsymbol{footnote}}
\footnotetext[1]{Equal contribution}
\begin{abstract}
Knowledge-based visual question answering (VQA) requires answering questions with external knowledge in addition to the content of images. One dataset that is mostly used in evaluating  knowledge-based VQA is OK-VQA, but it lacks a gold standard knowledge corpus for retrieval. Existing work leverage different knowledge bases (e.g., ConceptNet and Wikipedia) to obtain external knowledge. Because of varying knowledge bases, it is hard to fairly compare models' performance. To address this issue, we collect a natural language knowledge base that can be used for any VQA system.
Moreover,  we propose a Visual Retriever-Reader pipeline to approach knowledge-based VQA. The visual retriever aims to retrieve relevant knowledge, and the visual reader seeks to predict answers based on given knowledge.  We introduce various ways to retrieve knowledge using text and images and two reader styles: classification and extraction. Both the retriever and reader are trained with weak supervision. Our experimental results show that a good retriever can significantly improve the reader's performance on the OK-VQA challenge. The code and corpus are provided in \href{https://github.com/luomancs/retriever\_reader\_for\_okvqa.git}{this link}.
\end{abstract}

\section{Introduction}
\begin{figure}[t]
    \centering
    \includegraphics[width=0.97\linewidth]{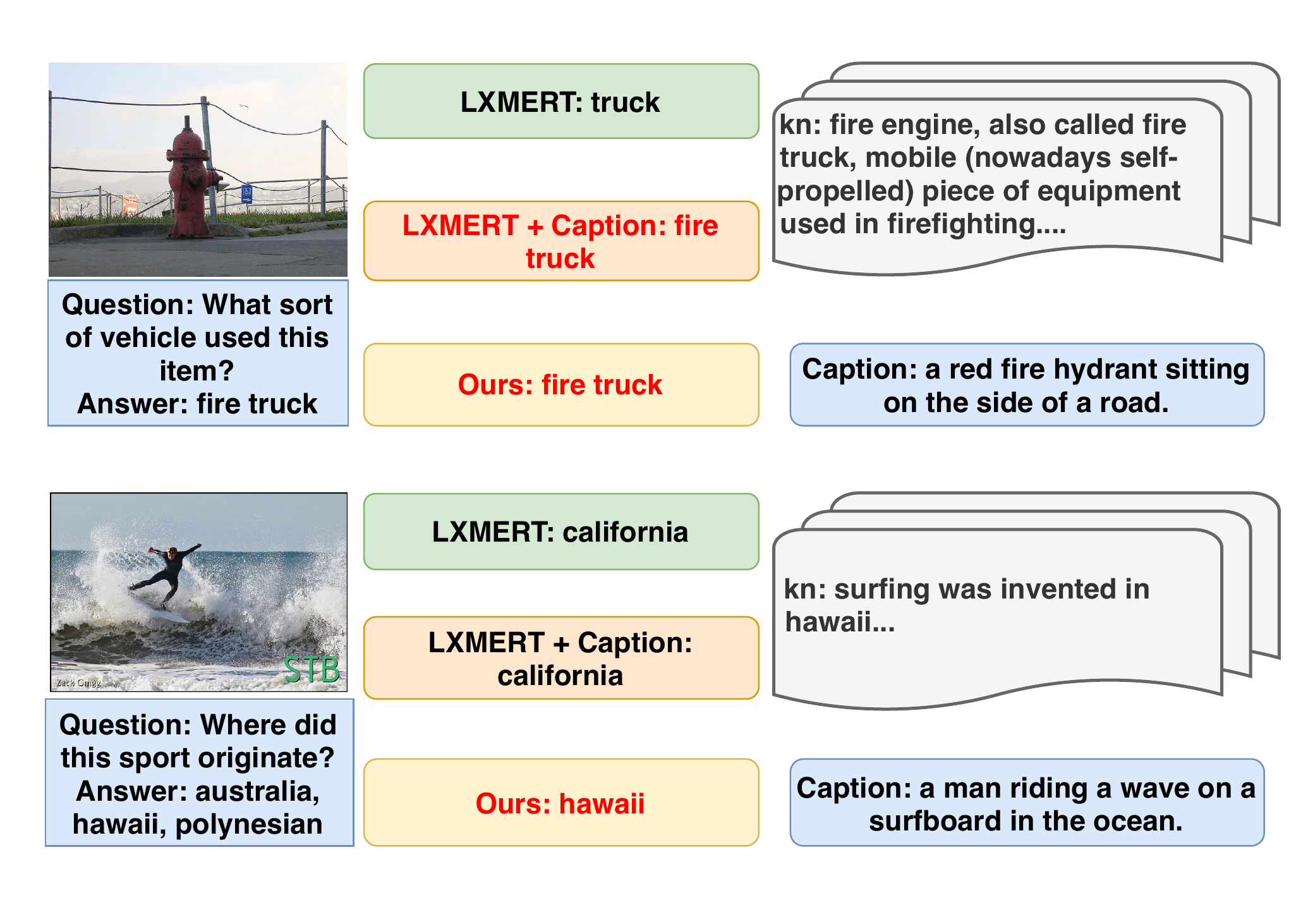}
    \caption{Two examples from OK-VQA: the middle column are predictions by two baselines and one by our proposed Visual-Retriever-Reader pipeline. The left column is relevant knowledge and the corresponding captioning of images.}
    \label{fig:example}
\end{figure}

Knowledge-based VQA is a challenging task, where knowledge present in an image is not sufficient to answer a question. It requires a method to seek external knowledge.  Figure \ref{fig:example} shows two examples from the OK-VQA benchmark~\citep{marino2019ok}, which is normally used to study knowledge-based VQA. In each of the two examples, external knowledge is needed to answer the question. For instance, in the first example, to identify the vehicle used in the item shown in the image (top-left), a system needs to first ground the referred item as a fire hydrant and then seek external knowledge presented top-right of the image. The challenge is to ground the referred object in the image and retrieve relevant knowledge where the answer is present. 

Although the OK-VQA benchmark encourages a VQA system to rely on external resources to answer the question, it does not provide a knowledge corpus for a QA system to use. As such, existing methods rely on different resources such as ConceptNet~\cite{speer2017conceptnet}, WordNet~\cite{miller1995wordnet}, and Wikidata~\cite{vrandevcic2014wikidata}, resulting in the following issues:

\begin{enumerate}[nosep,noitemsep,leftmargin=*]

\item It is difficult to fairly compare different VQA systems as it is unclear whether the difference in performance arises from differing model architectures or the different knowledge sources.

\item The different formats of the knowledge sources,  such as the structured  ConceptNet and the unstructured Wikipedia, demand different modules to retrieve knowledge, consequently making a knowledge-based VQA system complicated.

\item External resources like ConceptNet and WordNet have limitations. 
First, they only cover a limited amount of knowledge. For example,  ConceptNet provides only 34 relation types, and there is a vast amount of knowledge that is hard to be described by a relation in a knowledge graph, such as,  \textit{describe the logo of Apple Inc.}
Second, constructing a structured knowledge base requires heavy human annotation and is not available in every domain. Thus, it limits the application of a knowledge-based VQA system that relies on a structured knowledge base.
\end{enumerate}

Therefore, there is a need for a general and easy-to-use knowledge base. Motivated by this, we collect a knowledge corpus for the OK-VQA benchmark. Our corpus is automatically collected via Google Search\footnote{\url{https://developers.google.com/custom-search/v1/}} by using the training-split question and the corresponding answers, and we provide a training corpus with 112,724 knowledge sentences and a full testing corpus with 168,306 knowledge sentences. The knowledge corpus is in a uniform format, i.e., natural language. Thus, it is easy to use by other OK-VQA methods. As we will show in \S\ref{sec:experiment}, the knowledge base provides rich information to answer OK-VQA questions. 

\begin{figure*}[t]
    \centering
    \includegraphics[width=\linewidth]{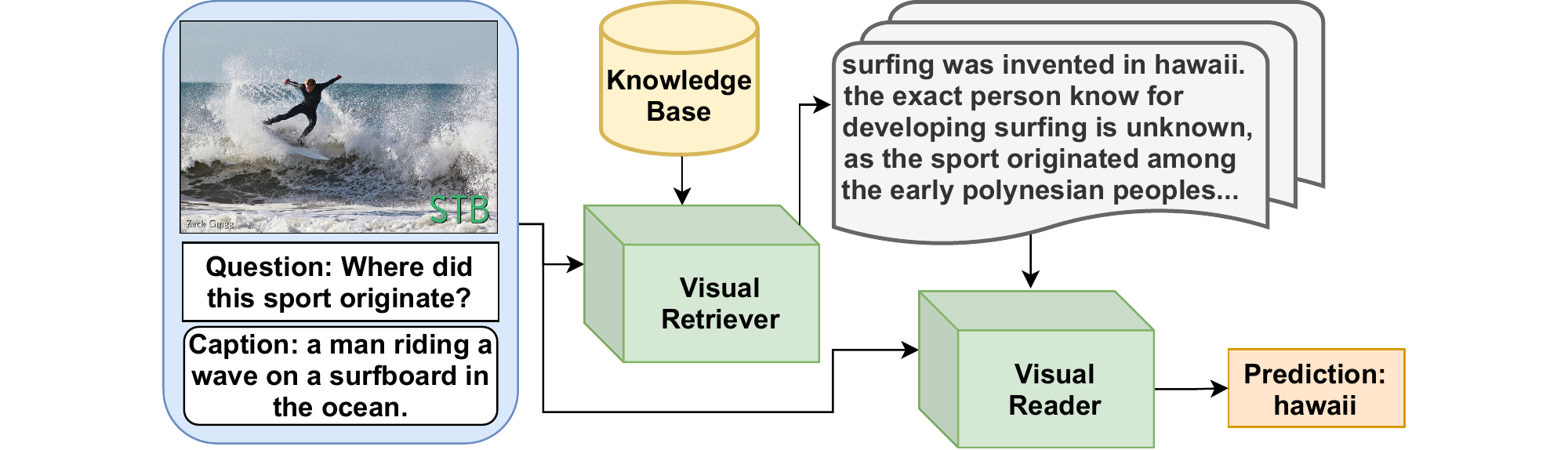}
    \caption{Visual Retriever Reader Pipeline: given an image and a question, a visual retriever is first to retrieve relevant knowledge, and then a visual reader is to predict an answer based on the given knowledge. }
    \label{fig:pipeline}
\end{figure*}

Utilizing the curated corpus, we further develop a weakly-supervised Visual-Retriever-Reader and evaluate it on the OK-VQA challenge. It consists of two stages, as seen in Figure \ref{fig:pipeline}. In the first stage, the visual retriever retrieves relevant knowledge from the corpus. In the second stage, the visual reader predicts an answer based on the given knowledge. Such a pipeline is well-studied in text-only open-domain QA ~\citep{chen2017reading, karpukhin2020dense}. We apply its principles to the multi-modal vision and language domain with novel adaptations. On the retriever side, we introduce visual information and evaluate a cross-modality model and a text-only caption-driven model (\S\ref{sec:retriever}). On the reader side, we build two visual readers, a classification and an extraction type, with both utilizing visual information (\S\ref{sec:reader}). We observe in \S\ref{sec:experiment}, our Visual-Retriever-Reader pipeline performs strongly on the OK-VQA challenge and establishes a new state-of-the-art.

Our experiments reveal multiple insights. First, we find that the image captions are very useful for both visual retriever and visual reader, which demonstrates the application of image captioning generator on knowledge-based VQA tasks. Second, a neural retriever has much better performance than a term-based retriever. This observation is quite interesting as in the NLP domain, typically, a term-based retriever (e.g., TF-IDF and BM25) is a hard-to-beat baseline~\cite{lee-etal-2019-latent,Lewis2020RetrievalAugmentedGF,ma-etal-2021-zero}, suggesting an essential role of neural retrievers in the vision-\&-language domain. Third, similar to the NLP domain, where a reader can perform well if the given knowledge contains relevant information, we discover that our visual reader has a significant leap when using noisy knowledge and high-quality knowledge. It motivates the need for developing a more efficient visual retriever for knowledge-based VQA tasks.

Our contributions are three folds. First, we build a general easy-to-use knowledge corpus for the OK-VQA benchmark, which makes model evaluation fair. Second, we propose a Visual-Retriever-Reader pipeline adapted from the NLP domain for the knowledge-based VQA task. Our model establishes a new state-of-the-art. Third, our experiments reveal several insights as mentioned above, and open a new research direction. 

\section{Related Work}

\paragraph{Knowledge-based VQA.} 

Many benchmarks have been proposed to facilitate the research in knowledge-based VQA. FVQA~\citep{wang2017fvqa} is a fact-based VQA dataset that provides image-question-answer-supporting fact tuples, where the supporting fact is a structured triple, e.g., $\langle$Cat, CapableOf,ClimbingTrees$\rangle$. KB-VQA~\citep{wang2015explicit} dataset consists of three types of questions: ``Visual" question answerable using the visual concept in an image, ``Common-sense" questions answerable by adults without looking for an external source, and ``KB-knowledge" questions requiring higher-level knowledge, explicit reasoning, and external resource. KVQA~\citep{shah2019kvqa} consists of questions requiring world knowledge of named entities in images. Specifically, the questions require multi-entities, multi-relation, multi-hop reasoning over Wikidata. KVQA is challenging, as linking the named entities in an image to the knowledge base is hard on a large scale. OK-VQA~\citep{marino2019ok} covers 11 types of knowledge than previous tasks, such as cooking and food, science and technology, plants and animals, etc. VLQA~\citep{sampat2020visuo} consists of data points of image-passage-question-answer, it is proposed recently to facilitate the research on jointly reasoning with both image and text. 


\paragraph{OK-VQA Systems.}
Out of the Box~\citep{narasimhan2018out} utilizes the Graph Convolution Networks~\citep{kipf2016semi} to reason on the knowledge graph (KG), wherein each node image and semantic embeddings are attached. Mucko~\citep{zhu2020mucko} goes a step further, reasoning on visual, fact, and semantic graphs separately, and uses cross-modal networks to aggregate them together.
ConceptBert~\citep{garderes2020conceptbert} combines the BERT-pretrained model~\citep{devlin2019bert} with KG. It encodes the KG using a transformer with a BERT embedding query. KRISP~\citep{marino2020krisp} involves a BERT-pretrained transformer model to make a better semantic understanding and utilize the implicit knowledge and reasons on a GCN model. 
Span-Selector ~\citep{jain2021select} extracts spans from the question to search most relative knowledge from Google, whereas MAVEx~\citep{wu2021multi} votes among textual and visual knowledge from Wikipedia, ConceptNet, and Google Image. Besides knowledge collection, knowledge alignment~\citep{shevchenko2021reasoning} also helps acquire a correct answer from knowledge.

\paragraph{Open-Domain Question Answering}
or ODQA tasks target collecting information from a large corpus to answer a question. The advanced reading comprehension model~\citep{chen2017reading,banerjee2019careful} split this complex task into two steps: a retriever selects some most relevant documents from a corpus to a question, and a reader produces answer according to the documents from retriever. Some previous work~\cite{kratzwald2018adaptive,lee2018ranking,das2019multi,wang2018r} train the end-to-end models to rerank in a closed set. Although these models are better at retrieval, they can hardly scale to larger corpora. Open-Retrieval Question Answering (ORQA)~\citep{lee2019latent} and Dense Passage Retriever (DPR)~\citep{karpukhin2020dense} constructed a dual-encoder architecture with BERT pre-trained model. This dense retrieval model shows a better performance than classic TF-IDF or BM25-based ODQA models on several natural language benchmarks.

\section{Knowledge Corpus Creation}\label{sec:knowledge}

\begin{figure*}[t]
    \centering
    \includegraphics[width=\linewidth]{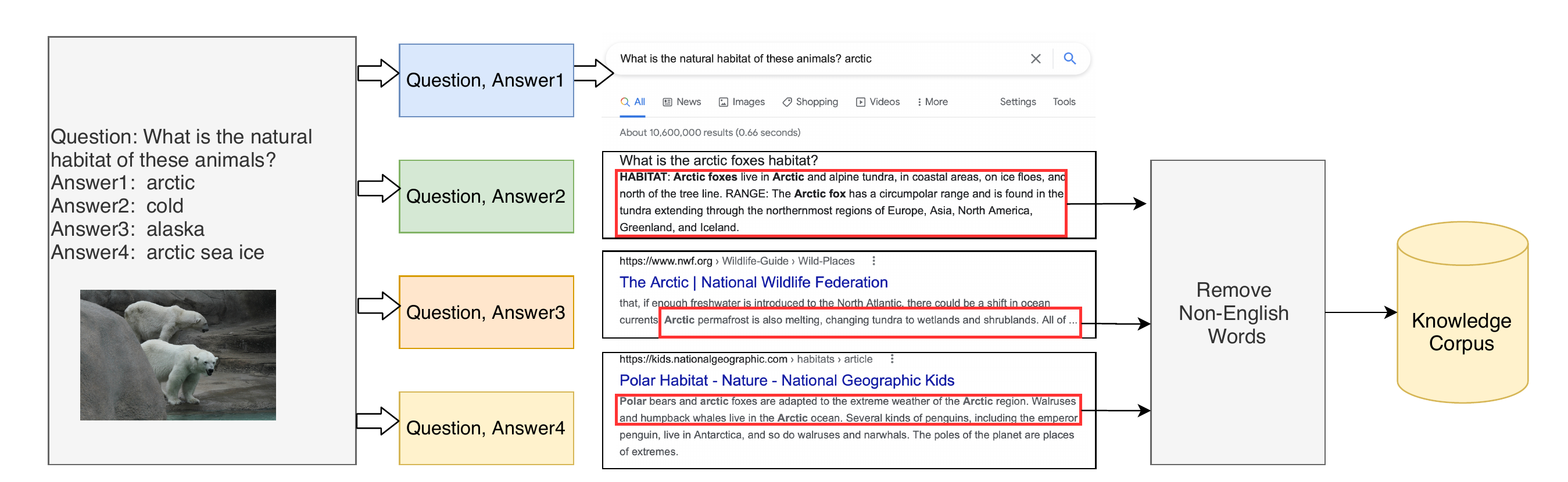}
    \caption{The overall process of Knowledge Corpus Creation. The question first combines the answers one by one to form a query, and then the query is sent to the Google Search API to retrieve the top \textit{10} webpages. The knowledge is obtained from the snippet with further processing. Finally, we integrate the knowledge into the corpus. As shown in the searching result page, the black boxes represent webpages, and red boxes represent snippets.}
    \label{fig:kn_collection}
\end{figure*}

The overall process of knowledge corpus creation (Figure \ref{fig:kn_collection}) consists of following four steps. 

\paragraph{Step 1: Query Preparation}
Based on the assumption that the knowledge used for answering training set questions can also help in testing, the OK-VQA training questions are used with their answers to collect related knowledge from a search engine. We concatenate each question with each answer to get a "Question, Answer" pair. For example, in Figure \ref{fig:kn_collection}, the question "What is the natural habitat of these animals?" has four answers, and each answer is attached to the question one by one to construct four queries.

\paragraph{Step 2: Google Search Webpage}
The generated queries are sent to Google Search API to obtain knowledge. As presented in Figure \ref{fig:kn_collection}, a good search result web page contains a title, a link, and a snippet that consists of multiple complete or incomplete sentences and shows the most relevant part to the query. The top \textit{ten} web pages with their snippets as the raw knowledge are chosen.

\paragraph{Step 3: Snippet Processing}
The snippets from Google searching results consist of multiple sentences, some are complete but some are not. One option is to split snippets into multiple sentences, but experimental result shows sentence-level knowledge is worse than snippet-level. Thus, we choose to use snippet as a knowledge. To address incomplete sentence issue, we find and grab the complete sentence present in the webpage.  After this pre-processing, ten snippet-knowledge from each "Question, Answer" query are selected.

\paragraph{Step 4: Knowledge Processing}
We first remove the duplicated data among each "Question, Answer" pair. Then long knowledge (more than 300 words) or short knowledge (less than ten words) are removed. Pycld2\footnote{\url{https://pypi.org/project/pycld2/}} is applied in this step to detect and remove the non-English part of each knowledge.  Each knowledge is assigned a unique ID and duplicate knowledge sentences are removed. We curate in total 112,724 knowledge sentences for the OK-VQA training set.

\section{Visual Retriever-Reader Pipeline}
We present our Visual Retriever-Reader pipeline for the OK-VQA challenge, where the visual retriever aims to retrieve relevant knowledge, and the visual reader aims to predict answers given knowledge sentences. This scheme has been widely used in NLP~\citep{chen-etal-2017-reading,karpukhin2020dense}. While previous work focuses on pure text-domain, we extend this to the visual domain with novel adaptation. 

\subsection{Retriever}\label{sec:retriever}
We introduce two styles of visual retriever: term-based and neural-network-based. In the neural style, we further introduce two variants. 
Following the convention, we use the standard terms in next subsection, for example, in \S\ref{sec:bm25}, we use \textit{documents} and in \S\ref{sec:dpr}, we use \textit{context}, both of them are \textit{knowledge} in our task.

\paragraph{Term-based Retriever.} \label{sec:bm25}
In BM25~\citep{Robertson2009ThePR}, each query and document is represented by sparse vectors in $d$ dimension space, where $d$ is the vocabulary size. Then the score of a query and a document is computed based on the inverse term's frequency. BM25 can only retrieve documents for a query in text format, but an image is a part of a query in our task. To tackle this issue, we first generate image captions using a caption generation model. Then we concatenate the question and the caption as a query and obtain a list of documents by BM25.

\begin{figure}[t]
    \centering
    \includegraphics[width=\linewidth]{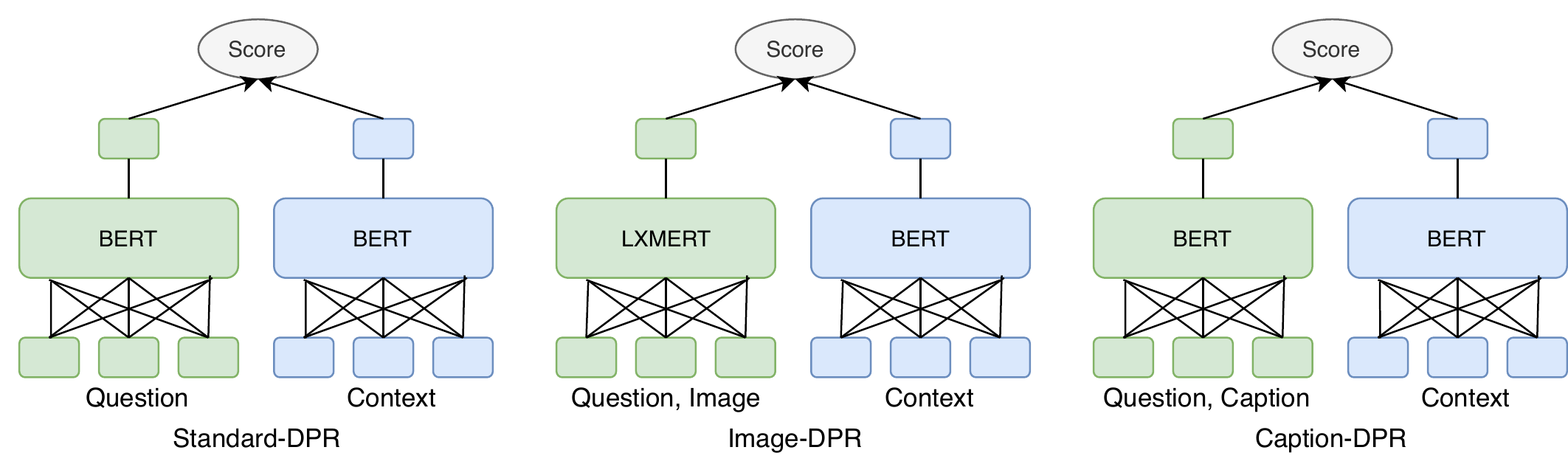}
    \caption{Comparison between standard DPR, Image-DPR and Caption-DPR: while the context encoder is the same for three models, in standard BERT(left), the question encoder only takes question as input, the Image-DPR(middle) takes both  question and image as input, the Caption-DPR (right) takes the question and the caption as input. }
    \label{fig:dpr}
\end{figure}

\paragraph{Neural Retriever.} \label{sec:dpr}
Unlike BM25, neural retrievers extract the dense representations for a query and a context from the neural model(s). We use DPR~\citep{karpukhin2020dense} as a neural retriever, which employs two BERT~\citep{devlin2019bert} models to encode the query and context respectfully, then applies inner-dot product to estimate the relevancy between a query and a context.  Similar to BM25, the DPR model considers the query in text format.  To adapt DPR in the visual domain, we propose two methods. \textit{Image-DPR}: we use LXMERT~\citep{Tan2019LXMERTLC} as the question encoder, which takes image and question as input and outputs a cross-modal representation.  \textit{Caption-DPR}: similar to the strategy we use in term-based retriever, we concatenate the question with the caption of an image as a query and use standard BERT as a query encoder to get the representation. In both \textit{Image-DPR} and \textit{Caption-DPR}, we use standard BERT as context encoder. Figure \ref{fig:dpr} shows the architectures of standard DPR, \textit{Image-DPR} and \textit{Caption-DPR}.
To train neural retriever, we use inner-dot product function to get the similarity score of relevant and irrelevant knowledge to a question, and optimize the negative log-likelihood of the relevant knowledge.

\subsection{Reader}\label{sec:reader}
\paragraph{Classification Reader (CReader).}
Current state-of-the-art VQA systems are classification models~\citep{Tan2019LXMERTLC,Li2019VisualBERTAS,gokhale2020vqa,gokhale2020mutant,banerjee-etal-2021-weaqa}, where a list of answer candidates are pre-defined (from the training set), i.e., a fixed answer vocabulary, then a model classifies one of the answers as the final prediction. We build a reader in this style but incorporate external knowledge. In particular, given a question, an image, and a piece of knowledge, we first concatenate the question with the knowledge and then apply a cross-modality  model to encode the text with the image and generate a cross-modal representation. We feed this representation to a Multiple Layer Perceptron (MLP) which finally predicts one of the pre-defined answers. We apply Cross-Entropy Loss to optimize the model. In this work, we use LXMERT~\citep{Tan2019LXMERTLC}, while any other cross-modality models like VisualBERT\citep{Li2019VisualBERTAS} can be adapted. 

\paragraph{Extraction Reader (EReader).}
The classification model fails to generalize to out-of-domain answers, i.e., questions whose answers are not in the pre-defined answer vocabulary. To tackle this issue, we use an extraction model which is adapted from machine reading comprehension model~\citep{chen-etal-2017-reading, karpukhin2020dense}.  The model extracts a span (i.e., a start token and an end token) from the knowledge to answer the question. The image caption is given to the model as well to incorporate the image information. We also inject a special word ``unanswerable'' before the caption so that the model can predict ``unanswerable'' if the given knowledge can not be relied on to answer the question. This strategy is helpful since the retrieved knowledge might be noisy. We use a  RoBERTa-large~\citep{liu2019roberta} as the text encoder, whose inputs are \{[SEP] question [SEP] [``unanswerable''], caption, knowledge [SEP]\}. Then each token representation is fed to two linear layers: one predicts a score for a token being the start token, and the other predicts a score for the end token. We apply the softmax function to get the probability of each token being a start and end token. The training objective is to maximize the probability of the ground truth start and end token. 
\subsection{Training and Inference}

\paragraph{Weak Supervision.} 
The retriever is trained using weak supervision, as the ground-truth knowledge context is unknown for a given question-image pair.
Particularly, given a query and an image, we assume that knowledge that contains any of the answers is relevant, and we use the in-batch negative samples~\citep{karpukhin2020dense} for training, i.e., in the training time, any relevant knowledge for other questions in the same batch are considered as irrelevant. 
For the reader, we use the same relevant knowledge as the retriever and when given such knowledge, the target is the answer. In addition, we use the same other collected knowledge which does not contain any answer as the irrelevant knowledge and in such a case, the reader should predict ``unanswerable'', a special word added to every knowledge. The reader may also be considered to be trained by weak supervision as the input knowledge is noisy, i.e., the assumed relevant knowledge is not guaranteed to be relevant. 

\paragraph{Inference Strategy.}\label{sec:strategy}
We use the retriever to retrieve K knowledge (the value and effects of K will be presented in \S\ref{sec:k}), and the reader predicts an answer based on each knowledge. We propose two strategies to predict the final answer. \textit{Highest-Score}: the answer which has the highest score is the final prediction. \textit{Highest-Frequency}: the answer which appears most frequently is the final prediction. 

\section{Evaluation}

\subsection{Retriever Evaluation}\label{sec:retriever-eval}
We evaluate the performance of a retriever based on Precision and Recall. The two metrics are based on the assumption that any retrieved knowledge that contains any of the answers annotated in the OK-VQA dataset is relevant. This assumption is because it is unknown which knowledge is relevant to a question-image pair. Therefore, the computation of Precision and Recall in our case is different from the traditional definition and illustrated as follows: 

\paragraph{Precision}
Precision reveals the proportion of retrieved knowledge that contains any of the answers to a question-image pair. Mean Precision is the mean of Precision of all question-image pairs. Mathematically, 
\begin{equation*}
P(Q, A, KN) = \frac{1}{K}\sum_{i=1}^{i=K} min (\sum_{\substack{j=1\\A_{j}\in KN_{i}}}^{j=M} 1, 1),
\end{equation*}
where $Q$ is a question, $KN$ is a list of retrieved knowledge, $A$ is a list of correct answers, $K$ is the number of $KN$, $M$ is the number of $A$.

\paragraph{Recall}
Recall reveals if at least one knowledge sentence in the retrieved Knowledge contains any answers to a question-image pair. Mean Recall is the mean of the Recall of all question-image pairs. Mathematically, 
\begin{equation*}
R(Q, A, KN) = min(\sum_{i=1}^{i=K}\sum_{\substack{j=1\\A_{j}\in KN_{i}}}^{j=M} 1, 1),
\end{equation*}
where the meaning of the symbols are the same described in Precision.

\subsection{Answer Evaluation.}

\paragraph{Original Evaluation}In OK-VQA, each image-question pair has five answers annotated by humans. To apply a similar evaluation as VQA~\citep{Agrawal2015VQAVQ}, OK-VQA counts per answer twice so that each image-question pair has ten answers, the same as VQA. The score is computed as follows. 

\begin{equation*}
score(A) = min(\frac{\# \textrm{human that said A}}{3}, 1) 
\end{equation*}

We use the above equation to compute the score of each answer for training and testing.

\paragraph{Open Domain Evaluation} 
\citet{luo-etal-2021-just} propose an evaluation method for VQA, especially mitigating the issues of Synonym/Hypernym and Singular/Plural. Here, we adapt their method to the open domain setting, where Sentence Textual Entailment (STE) is used to find semantically similar answers. In STE, given a premise and a hypothesis, a score is generated to indicate whether a premise entails the hypothesis. In our case, a premise is a sentence that contains a gold answer, and a hypothesis is the same sentence while the gold answer is replaced by a predicted answer given by a model. Suppose a high STE score\footnote{We only credit those predicted answers which obtain an STE score higher than 0.5.} is generated for such a pair of premise and hypothesis. In that case, it implies that the predicted answer is semantically close to the gold answer and thus deserves a partial score. We provide the detailed steps of evaluation in Appendix \ref{apd:od_eval}.

Mathematically, the open-domain accuracy is given by the follows, 
\begin{equation*}
\begin{split}
   & \text{S}_j(Q, I, a_j, a') = \frac{1}{| \text{P}_{a_j}|}\sum_{g_i \in \text{P}_{a_j}}\text{E}(a_j,a', g_i) \\
   & \text{S}'(Q, I, a') = \argmax_{a_j\in Ans}\text{S}_j(Q, I, a_j, a) \\
   & \;\;\;\;\;\;\;\;\;\;\;\;\;\;\;\;\;\;\; \times \text{S}(Q, I, a_j)
   \end{split}
\end{equation*}
where $a'$ is a predicted answer, $Ans$ is a set of ground truth answer of a pair of question(Q) and image(I), $\text{P}_{a_j}$ is a set of sentences with placeholder,  $\text{E}(a_j,a', g_i)$ is the entailment score of a premise (sentence $g_i$ grounded by gold answer $a_j$) and hypothesis (sentence $g_i$ grounded by predicted answer $a'$), and S$(Q, I, a_j)$ is the ground truth score of $a_j$ which has highest entailment score with predicted answer $a'$.

The main difference between our evaluation and \citep{luo-etal-2021-just} is that in their evaluation, they extend each answer with a set of alternative answers which are selected from the list of pre-defined answers in the training set. While in our setting, we remove this step since in the open domain setting, many semantically similar answers can be found in the open corpus which is not necessary in the training set (see examples in Appendix \ref{apd:examples_odeval}).

\section{Experiments and Results}\label{sec:experiment}
\subsection{Baselines}
We use a state-of-the-art vision-language model, LXMERT ~\citep{Tan2019LXMERTLC}, as the baselines and apply Captioning and Optical Character Recognition (OCR) results to the OK-VQA dataset to the original LXMERT model.

\paragraph{LXMERT}
LXMERT is a BERT-based cross-modality model pretrained on five different VQA datasets:  MS COCO~\citep{lin2014microsoft}, Visual  Genome~\citep{krishna2017visual}, VQA  v2.0~\citep{antol2015vqa}, GQA balanced version~\citep{hudson2019gqa} and VG-QA~\citep{zhu2016visual7w}. We fine-tune LXMERT on OK-VQA and surprisingly find that LXMERT ranks higher than most of the SOTA models, for which reason we set LXMERT as our baseline model.

\paragraph{LXMERT with OCR} 
The OCR technique captures the textual contents from the image and transfers them into characters. Here we use Google Vision API\footnote{\url{https://cloud.google.com/vision/}} to extract the texts from images. After the noise deduction step filtering all non-English words, we attach the OCR results after the question and then sent them into the LXMERT model. Our experiment shows that the OCR result helps to address the OK-VQA task.

\paragraph{LXMERT with Captioning}
Similar to OCR, we also experiment with adding captioning when training the LXMERT model. The captions are generated by the advanced model Oscar~\cite{li2020oscar} and attached to each question when sent into the LXMERT model. Our result shows that captioning improves the performance of the LXMERT model, and therefore, we put the LXMERT with captioning as a baseline as well.

\begin{table}[]
    \centering
    \resizebox{\linewidth}{!}{
    \begin{tabular}{lccc}
    \toprule
    {\bf Method}  & {\bf Knowledge Src.} & {\bf Acc.} & {\bf Open Acc.}\\
    
    \toprule
    {\bf Existing Method} & ~ & ~ \\ 
    \hline
    KRISP~\citep{marino2020krisp} & W \& C & 32.3 & -\\
    ConceptBert~\citep{garderes2020conceptbert} & C & 33.7 & - \\
    MAVEx~\citep{wu2021multi} & W \& C \& GI & 38.7 & -\\ 
    \hline
    {\bf Baselines} & ~ & ~ & ~\\
    \hline
    LXMERT (without pretraining) & - & 18.9 & 25.5\\
    LXMERT & - & 36.2 & 42.6 \\
    LXMERT + OCR & - &  37.2 & 42.2\\
    LXMERT + Caption & - & 37.8 & 45.6 \\
    LXMERT + OCR + Caption & - & 37.2 & 44.5 \\
    \hline
    {\bf Visual Retriever-Reader} & ~ &~ & ~\\
    \hline
    BM25 + CReader & GS & 35.13 & 43.8\\
    BM25 + EReader  & GS & 32.10 & 40.6\\
    Image-DPR  + CReader & GS  & 34.64 & 43.2\\
    Image-DPR + EReader & GS & 33.95 & 41.7\\
    Caption-DPR  + CReader & GS & 36.78 & 43.4\\
    Caption-DPR + EReader & GS & {\bf 39.20} & {\bf 47.3}\\
    Caption-DPR + EReader $^\dagger$ & GS& 59.22 & 66.6\\
    \toprule
    \end{tabular}
    }
    \caption{Performance on the OK-VQA Test-split. Our model outperforms existing methods.  $\dagger$ means given oracle knowledge to the reader. GS-Google Search (Training Corpus). W-Wikipedia, C-ConceptNet, GI-Google Image, Acc-Accuracy.}
    \label{tab:main-result}
\end{table}

\begin{table*}[ht!]
    \centering
    
    \resizebox{\textwidth}{!}{
    \begin{tabular}{lcccccccccccccc}
    \toprule
    \multirow{3}{*}{\bf Model}   &\multicolumn{12}{c}{\bf \# of Retrieved Knowledge }\\
    \cmidrule(lr){2-13}
    ~& \multicolumn{2}{c}{1} & \multicolumn{2}{c}{5}   &\multicolumn{2}{c}{10}  & \multicolumn{2}{c}{20}  & \multicolumn{2}{c}{50}  & \multicolumn{2}{c}{80} & \multicolumn{2}{c}{100} \\
    \cmidrule(lr){2-3}\cmidrule(lr){4-5}\cmidrule(lr){6-7}\cmidrule(lr){8-9}\cmidrule(lr){10-11}\cmidrule(lr){12-13}\cmidrule(lr){14-15}
    ~ &  P* & R* &  P* & R* &  P* & R* &  P* & R* &  P* & R* &  P* & R* &  P* & R* \\
    \toprule
    BM25 & 37.63 & 37.63 &  35.21 & 56.72 &  34.03 & 67.02 & 32.62 & 75.90 &  29.99 & 84.56 & 28.46 & 88.21 & 27.69 & 89.91 \\
    Image-DPR &  33.04 & 33.04 & 31.80 & 62.52 &  31.09 & 73.96 &  30.25 & 83.04 &  28.55 & 90.84 &  27.40 & 93.80 & 26.75 & 94.67 \\
    Caption-DPR & {\bf 41.62} & {\bf 41.62} &  {\bf 39.42} & {\bf 71.52} &  {\bf 37.94} & {\bf 81.51} &  {\bf 36.10} & {\bf 88.57} &  {\bf 32.94} & {\bf 94.13} & {\bf 31.05} &  {\bf 96.20}  &{\bf 30.01} & {\bf 96.95}\\
    \toprule
    \end{tabular}
    }
    \caption{
    Evaluation of three proposed visual retrievers on Precision and Recall: 
    Caption-DPR achieves the highest  Precision and Recall on all number of retrieved knowledge. 
    We have a * marker on the {\bf P}recision and {\bf R}ecall to distinguish from traditional Precision and Recall as illustrated in \S\ref{sec:retriever-eval}. }
    \label{tab:retriever-eval}
\end{table*}

\subsection{Main Results}

Table \ref{tab:main-result} shows that our best model based on Caption-DPR and EReader outperforms previous methods and establishes the new state-of-the-art result on the OK-VQA challenge. Interestingly, the LXMERT baseline without utilizing any knowledge achieves better performance than KRISP~\citep{marino2020krisp} and ConceptBert~\citep{garderes2020conceptbert} which leverage external knowledge. Incorporating OCR and captioning further improve the baseline accuracy by 1\% and 1.6\%, respectively. 

Among different variations of Visual Retriever-Reader, the best combination is Caption-DPR and CReader when the retrieved knowledge size is 80. We evaluate retrievers' performance based on Precision and Recall. Table \ref{tab:retriever-eval} shows that Caption-DPR consistently outperforms BM25 and Caption-DPR on the various number of retrieved knowledge. It is interesting to see that Caption-DPR outperforms BM25 significantly since BM25 is a hard-to-beat baseline in open-domain QA~\citep{lee-etal-2019-latent,Lewis2020RetrievalAugmentedGF,ma-etal-2021-zero}. It indicates that neural retriever has better application than term-based retrieval methods in the vision domain. 

We also present the results obtained by open domain evaluation. First, we observe that the open domain evaluation is correlated with the original accuracy evaluation, i.e., the higher the original accuracy is, the higher the open domain accuracy is. Second, by open domain evaluation, the score is higher than before as some semantic similar answers get credits. We present some examples in Appendix \ref{apd:examples_odeval}.

\begin{table*}[]
    \centering
    \begin{tabular}{lccccccc}
    \toprule
    \multirow{2}{*}{\bf Model}  &\multicolumn{7}{c}{\bf \# of Retrieved Knowledge }\\
    \cmidrule(lr){2-8}
    ~&  1 & 5  &10 & 20 & 50 & 80 &  100\\
    \toprule
    
    BM25 &+6.00  & +6.28  &+4.88 & +4.32 & +3.83  & +3.17 & +2.56\\
     
    Image-DPR&+2.24 & +2.60 & +2.93 &  +2.29 & +1.83  &+1.29 &+1.25 \\ 
    
    Caption-DPR & +8.88 & +8.88  &+7.04 & +4.65 & +2.98  & +2.23 & +1.88\\
    \toprule
    \end{tabular}
    \caption{Recall increases when the Caption-DPR method retrieves knowledge from a complete knowledge corpus created using train and test questions.}
    \label{tab:corpus}
\end{table*}

\section{Analysis}\label{sec:analysis}

\paragraph{Effects of the Quality of Knowledge.}
A common observation in open-domain question answering in NLP is that the reader can perform well if the given knowledge is good to answer a question. Here, we are interested to see if this also holds for our reader. Specifically, before we feed the retrieved knowledge to the reader, we remove knowledge that does not contain any answer, then we send the remaining knowledge to the reader. The last row in Table \ref{tab:main-result} shows that our reader can perform much better if the quality of the knowledge is good, suggesting that a more efficient cross-modality retriever is needed.  

\begin{figure}[t]
    \centering
    \includegraphics[width=0.75\linewidth]{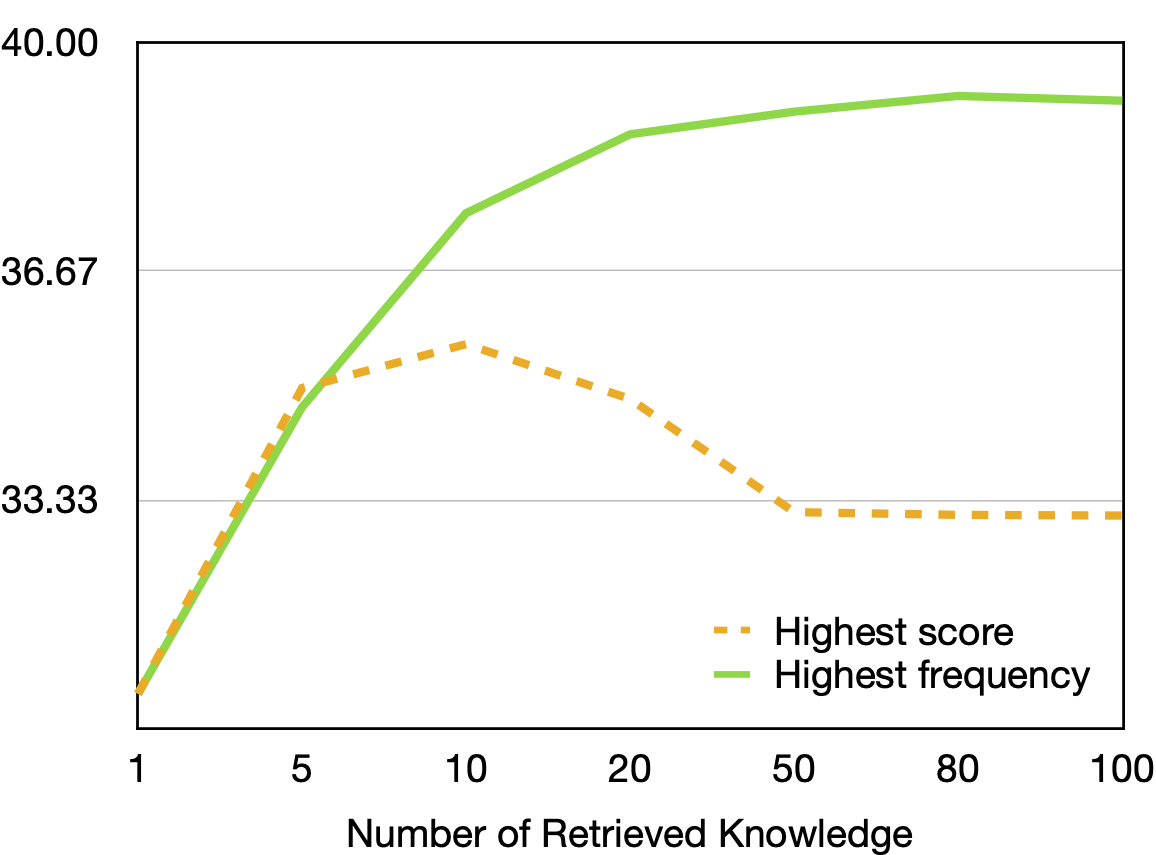}
    \caption{
    Highest-Score Strategy: Performance of EReader decreases when the knowledge number increase and the best is at 5.  Highest-Frequency strategies: Performance of EReader increase when the knowledge number increase and the best is at 80.
    }
    \label{fig:okvqa-kn-size}
\end{figure}

\paragraph{Effects of Size of Retrieving Knowledge and Prediction Strategy.}\label{sec:k} The performance of reader is directly affected by the size of retrieved knowledge. A more extensive knowledge set is more likely to include the relevant knowledge to answer the question yet along with more distracting knowledge. In contrast, a small set might exclude relevant knowledge but with fewer distracting knowledge. We use Caption-DPR to retrieve the different number of pieces of knowledge and use the EReader to predict an answer given the different number of pieces of knowledge. We compare the effects on two prediction strategies mentioned in \S\ref{sec:strategy}.  Figure \ref{fig:okvqa-kn-size} shows the comparison, and we have the following observations. First, when the knowledge size is small (equal or less than 5), the Highest-Score strategy is better than the Highest-Frequency; on the other hand, when the knowledge size is large, the Highest-Frequency strategy performs better than the Highest-Score strategy. Second, for the Highest-Score strategy, the size of 5 is the best, and increasing the knowledge size reduces the performance. Third, for the Highest-Frequency strategy, when the size equal to 80, it yields the best performance. To summarize, if one uses a small set of knowledge, then Highest-Frequency negatively impacts the accuracy and the Highest-Score strategy is preferable. If one uses a larger corpus of knowledge, the Highest-Frequency strategy can achieve higher accuracy. 

\begin{figure}[t]
    \centering
    \includegraphics[width=0.75\linewidth]{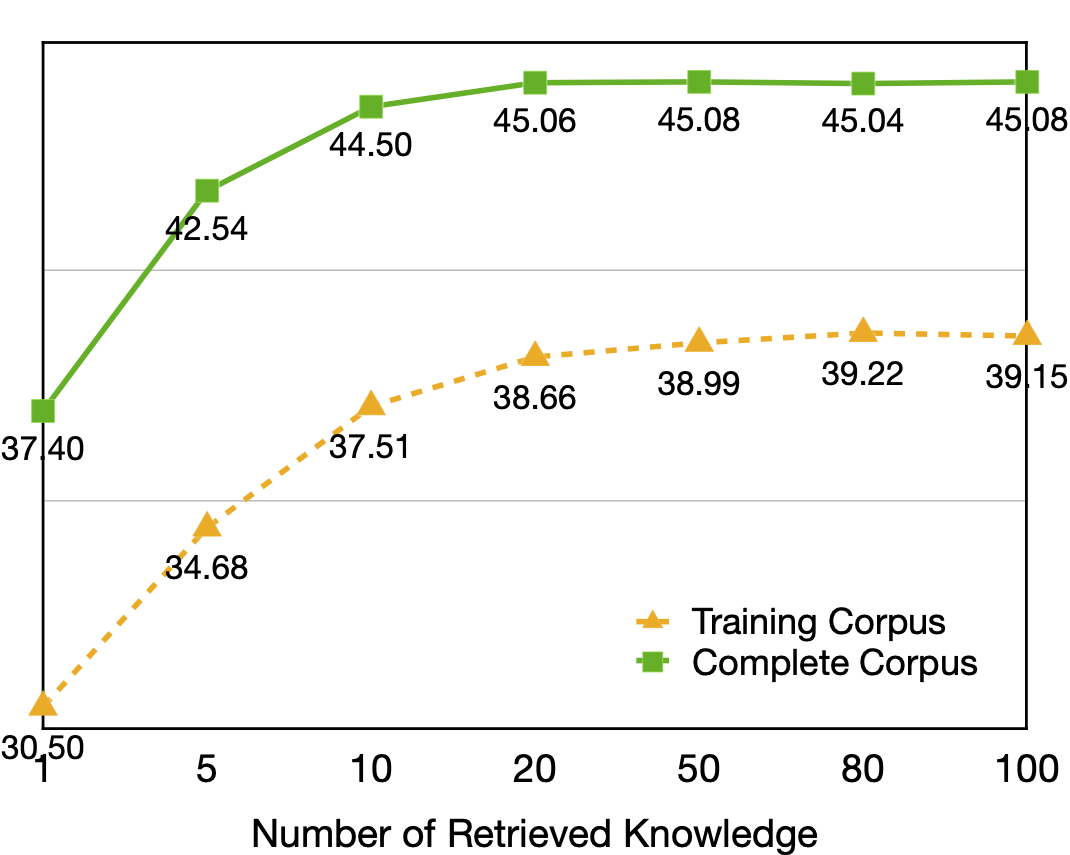}
    \caption{EReader achieves significant improvement when using knowledge retrieved from complete corpus compared to knowledge from training corpus.}
    \label{fig:okvqa-corpus}
\end{figure}

\paragraph {Effects of Completeness of Corpus.}
So far, when we test the model performance, we use the knowledge corpus collected only by training questions. However, if the entire training corpus does not include relevant knowledge to testing questions, our model is under-evaluated because of the incompleteness of the knowledge corpus. To fairly see how our model performs when the knowledge corpus is complete, we use the same knowledge collection method described in \S\ref{sec:knowledge} to collect knowledge for testing questions. Then we combine the training and testing knowledge as a complete corpus, which increases the corpus size from 112,724 to 168,306. We use Caption-DPR to retrieve knowledge from the complete corpus and ask EReader to predict answers based on these pieces of knowledge. Table \ref{tab:corpus} shows the increase of recall. As we expected, a complete corpus is helpful for Caption-DPR even though the corpus size increased, thus yields better performance of EReader.  Figure \ref{fig:okvqa-corpus} compares the accuracy of EReader using knowledge retrieved from two corpora. EReader consistently achieves higher performance using the knowledge retrieved from complete corpus, where the biggest gain of 7.86\%  is achieved when using five knowledge. 
We further clean up the corpus following similar steps in \citep{raffel2019exploring}, and 1\% of the knowledge got removed from the initial ones. We provide the details in Appendix \ref{apd:clean_corpus}.

\section{Discussion} 




\paragraph{Training Corpus Bias}
One potential concern of our knowledge corpus is that the training corpus might tend to bias to the training set, i.e., the training corpus includes knowledge for the training set and excludes knowledge for some testing set. To alleviate such concern, we analyze the training and testing sets in OKVQA and find that 74.69\% of testing answers overlap with the training answers, which indicates that a large portion of common knowledge is shared by training and testing sets. To further complement the training corpus, we also provide the entire corpus for testing (discussed in \S\ref{sec:analysis}), which also includes relevant knowledge for the testing set. The entire corpus is larger than the training corpus and includes prior unseen knowledge. Thus, such testing corpus can evaluate the generalization ability of a retriever, which is an essential skill of any AI system~\citep{mishra2021natural}.

\paragraph{Extension.}
Although our pipeline is evaluated on the OK-VQA benchmark, it is generic and can be adapted for other knowledge-based question answering tasks such as FVQA~\citep{wang2017fvqa}, KB-VQA~\citep{wang2015explicit}, and KVQA~\citep{shah2019kvqa}. For example, in KVQA, we can first collect a named-entity knowledge corpus by the proposed knowledge collection approach and then apply our Visual-Retriever-Reader pipeline.  
It should be noted that our proposed Extractive reader is a more challenging problem as classification models tend to learn correlation between output classes (answers)~\cite{agarwal2020towards} and input image and question. In contrast, the extractive reader extracts answer-spans which we exactly match with targets (answers).

\section{Conclusion}
This paper collects an easy-to-use free-form natural language knowledge corpus for VQA tasks with external knowledge. A weakly-supervised Visual Retriever-Reader Pipeline, where the retriever introduces dense representation, and the reader contains classification and extraction two styles, is also evaluated. The Visual Retriever-Reader Pipeline has been evaluated on the OK-VQA challenge benchmark and has established a new state-of-the-art performance. Further analysis reveals that good knowledge from the retriever makes vital progress in predicting the correct answer. Besides, the captioning and the neural retriever can both significantly improve the QA system's performance.

\nocite{Ando2005,borschinger-johnson-2011-particle,andrew2007scalable,rasooli-tetrault-2015,goodman-etal-2016-noise,harper-2014-learning}

\section*{Acknowledgements}

The authors acknowledge support from the NSF grant 1816039, DARPA grant W911NF2020006, DARPA grant FA875019C0003, and ONR award N00014-20-1-2332; and thank the reviewers for their feedback.

\section*{Ethical Considerations} 
All the knowledge base data is fully automatically collected through the website source open to public. We select several sentences from these pages of passages for non-commercial use, which will not violate the intellectual property and privacy rights. The whole dataset is aimed to address the external knowledge source for Knowledge-based QA, and thus each piece contains some commonsense or encyclopedic knowledge. To ensure that the sentences are properly excerpted from the source page, we also visit the source website to extract the complete sentences. (see Step 3 of \S\ref{sec:knowledge}). Our experiment (\S\ref{sec:experiment}) also shows that the collected knowledge is helpful to answer the OK-VQA questions.


\bibliography{anthology,custom}
\bibliographystyle{acl_natbib}

\appendix

\section{Open Domain Evaluation} \label{apd:od_eval}
\begin{figure*}[t]
    \centering
    \includegraphics[width=\linewidth]{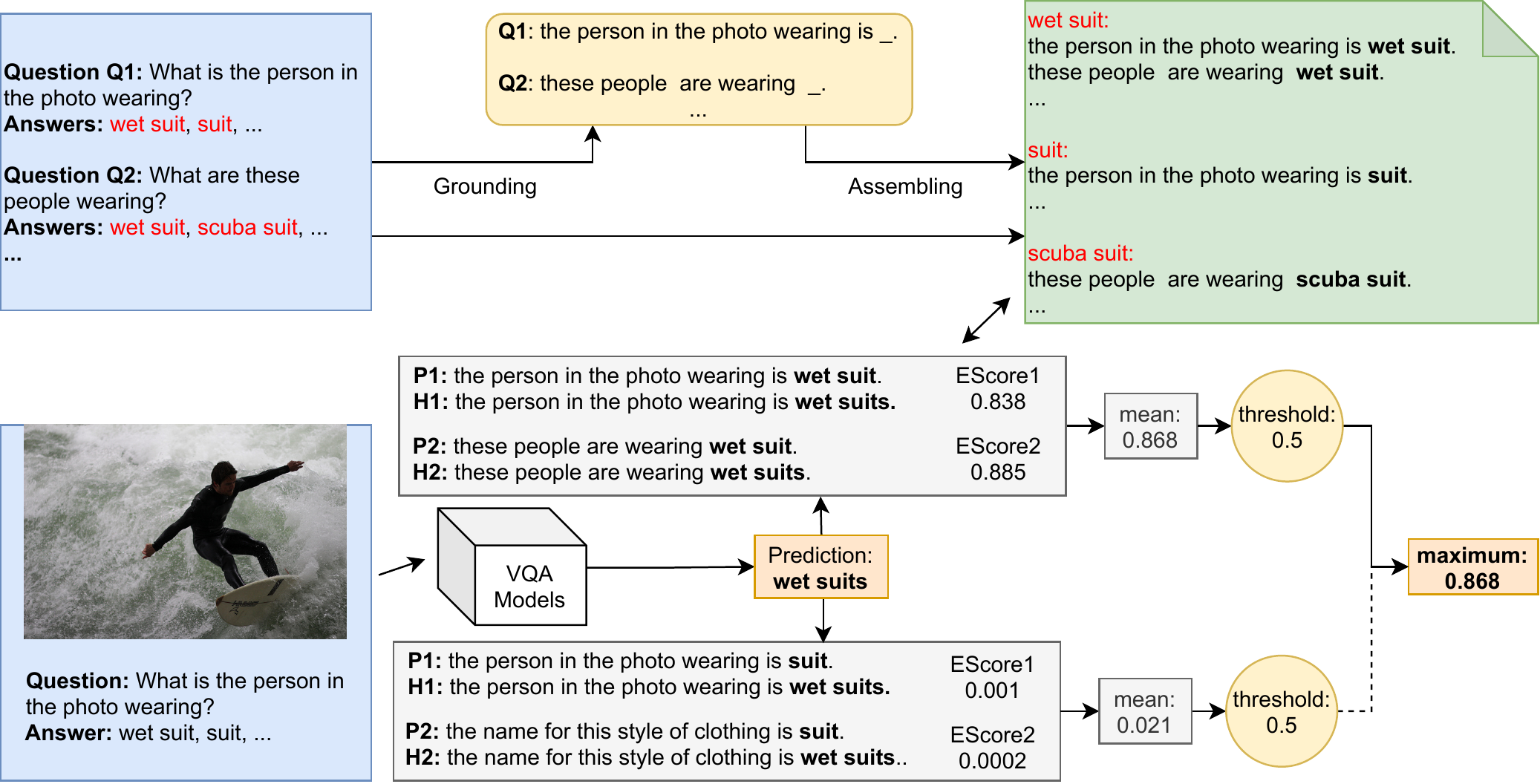}
    \caption{This example calculates the entailment score of provided answer ``wet suit'' and our prediction ``wet suits''. We first ground all questions into statements with a reserved position ``\_'' for the answer. Then, we congregate all the grounded statements by the provided answer. We replace the ``\_'' with the provided and predicted answer separately as the premise and hypothesis to get the entailment score. The entailment score of a provided answer and a prediction is calculated as the mean of all the entailment scores under that answer in the assembling list. We take 0.5 as the threshold, and use the maximum as the final entailment score.}
    \label{fig:entailment}
\end{figure*}

Particularly, our evaluating contains three phases: \textit{Grounding} that apply each answer and prediction to the question to ground it as a statement; \textit{Assembling} that rearrange the statements by answers; and \textit{Entailment} that calculate the similarity of the different grounded sentences. Then the final score is calculated according to the STE results. Figure \ref{fig:entailment} gives an overview of the open-domain evaluation.
Considering that the Extraction Reader predicts an answer within the open domain, probably resulting in the generated phrases not showing up in the answer field,

\begin{figure*}[t]
    \centering
    \includegraphics[width=\linewidth]{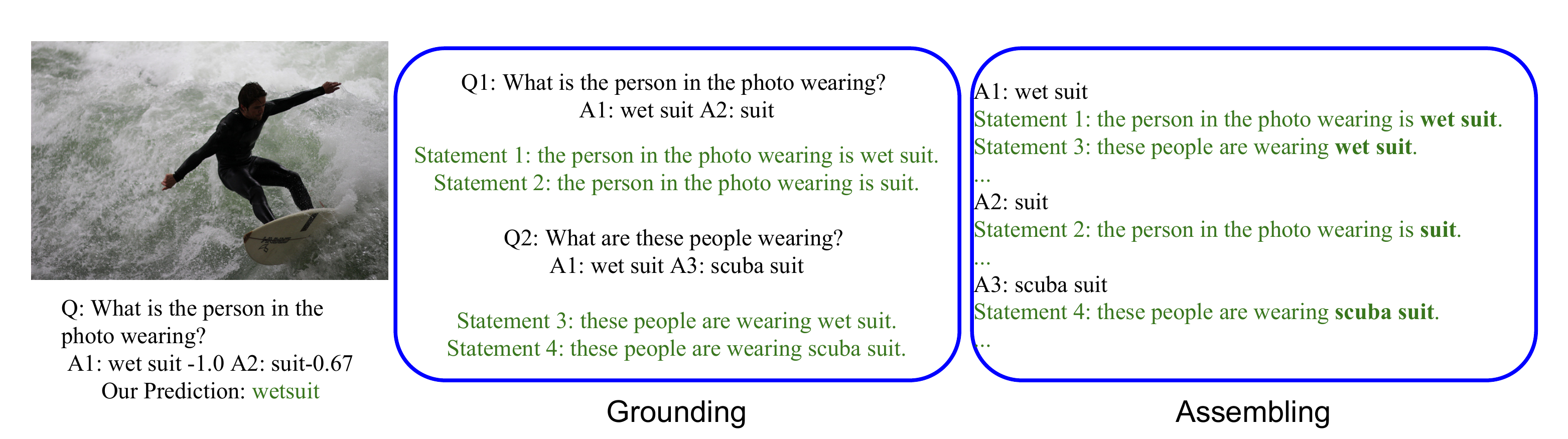}
    \caption{Example of Grounding and Assembling Step in Open-Domain Evaluation. }
    \label{fig:ground}
\end{figure*}

\subparagraph{Grounding}
In the grounding phase, we convert a question to a statement using the answers and predictions. Since a good prediction should be of the similar semantic meaning as the answers, we assume that for one question, every answer and prediction acts a same role in the grounded statement, and thus we ground the question with a reserved position for any answer to fill in. For example, the original question ``Who invented this device?'' is grounded to ``\_ invented this device.'', where ``\_'' can be any of the answers to this question. An example for grounding is shown in Figure \ref{fig:ground}.

To achieve this, a simple sentence role labeling work is applied to the questions to detect different elements in the sentence (e.g. question word, object, subject, auxiliary word, etc.) After settling the role of elements, the question is then re-ordered to accord with the word order of declarative sentences.
We apply the above method to the wh-questions and choice questions, which in total cover the 98.6\% of questions and 98.9\% of unique answers. Table \ref{tab:ground} shows some examples of grounded sentences.

\begin{table*}[]
    \centering
    \resizebox{0.97\linewidth}{!}{
    \begin{tabular}{c c}
        
        \toprule
        \textbf{Original Question} & \textbf{Grounded Statement} \\ \hline
        What is this type of blanket called? & this type of blanket is called \_. \\
        What is the name of the board he is on? & the name of the board he is on is \_. \\
        The food in the photo contains which healthy vitamins? & The food in the photo contains \_ healthy vitamins. \\
       Is this bathroom high or low end? & this bathroom is \_. \\
       Why is the cow going to the water? & the cow  is going to the water  because of \_. \\
        \toprule
    \end{tabular}
    }
    \caption{Examples for some grounded sentences where the hypothesis gets score over the threshold.}
    \label{tab:ground}
\end{table*}

\subparagraph{Assembling}
In grounding step, the statements are gathered by question. We re-arrange the these grounded statements ordered by the provided answers for the further processing. Figure \ref{fig:ground} provides an example for this assembling step.

\subparagraph{Entailment}
The grounded sentences are then sent to the Natural Language Inference (NLI) model \footnote{\url{https://github.com/allenai/allennlp-models/tree/v1.0.0.rc2/training_config/nli}}. NLI is used widely in the NLP tasks to check whether the hypothesis can be entailed from the given premise, and here we use NLI to check whether the correct answers and the predicted answer are semantically same. To compare between a provided answer and a predicted answer, we first list all grounded statements that use the provided answer as a correct answer. Then, for each of these statements, we fill the reserved position with the provided answer as the premise, and our prediction is the hypothesis, and calculate the entailment score. We use the arithmetic mean of these scores as the final entailment score.

The threshold is set to be 0.5. We also skip the choice questions and the questions with numbers as answers, since, with only grounded statements provided, it is hard to tell whether the two numbers or two choices are similar. For each question with multiple answers, we pick the highest entailment score as the similarity score.

\begin{figure*}[t]
    \centering
    \includegraphics[width=\linewidth]{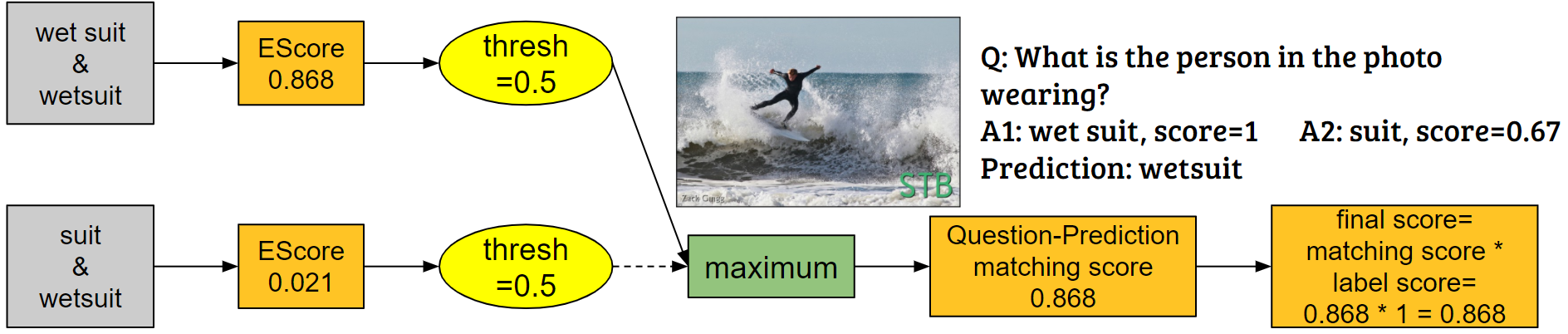}
    \caption{Example of Entailment Step in Open-Domain Evaluation. }
    \label{fig:nli}
\end{figure*}

Figure \ref{fig:nli} shows the steps acquiring the entailment score and calculating the final score for a predicted answer.

\section{Examples of Open Domain Evaluation} \label{apd:examples_odeval}

Here, we show four examples such that the predicted answer given by our model is in fact semantically close to one of the ground truth answer. Such predictions get 0 score given the original accuracy evaluation, but get reasonable score by our proposed open domain evaluation.  

\begin{figure*}[ht!]

\centering
\subfigure{
    \begin{minipage}[t]{0.48\linewidth}
    \includegraphics[width=0.95\linewidth]{open1}
    \small
    \begin{tabular}{l}
    {\bf Question}: Is this a room for a boy or a girl?\\
    {\bf Ground-Truth Answer}: girl : 1.0\\
    {\bf Our Prediction: girls}\\
    {\bf Original Score}: 0\\
    {\bf Open-Evaluation Score}: 0.88
    \end{tabular}
    \end{minipage}
}
\subfigure{
    \begin{minipage}[t]{0.48\linewidth}
    \includegraphics[width=0.95\linewidth]{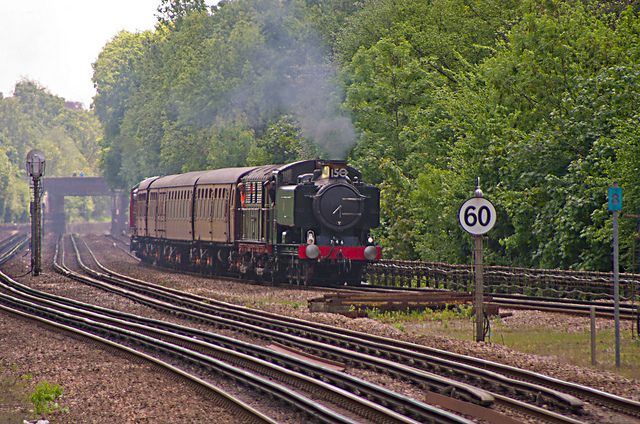}
    \small
    \begin{tabular}{l}
    {\bf Question}: Name the model of train shown in this picture?\\
    {\bf Ground-Truth Answer}: steam locomotive : 1.0;\\e2 class steam locomotive : 0.67; steam engine : 0.67\\
    {\bf Our Prediction}: locomotive\\
    {\bf Original Score}: 0\\
    {\bf Open-Evaluation Score}: 0.63
    \end{tabular}
    \end{minipage}
}
\subfigure{
    \begin{minipage}[t]{0.48\linewidth}
    \includegraphics[width=0.95\linewidth]{open3}
    \small
    \begin{tabular}{l}
    {\bf Question}: What is needed to use this vehicle?\\
    {\bf Ground-Truth Answer}: license : 1.0; gasoline : 0.67\\
    {\bf Our Prediction}: fuel\\
    {\bf Original Score}: 0\\
    {\bf Open-Evaluation Score}: 0.61
    \end{tabular}
    \end{minipage}
}
\subfigure{
    \begin{minipage}[t]{0.48\linewidth}
    \includegraphics[width=0.95\linewidth]{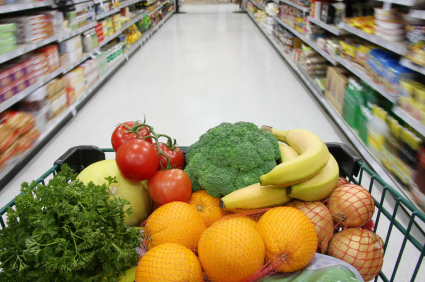}
    \small
    \begin{tabular}{l}
    {\bf Question}: What place of business is this?\\
    {\bf Ground-Truth Answer}: grocery store : 1.0; \\supermarket : 0.67\\
    {\bf Our Prediction}: grocery\\
    {\bf Original Score}: 0\\
    {\bf Open-Evaluation Score}: 0.86
    \end{tabular}
    \end{minipage}
}

\caption{\small Examples for Open Domain Evaluation.}
\label{fig:open-eg}
\end{figure*}

\section{Training Setup}
Our neural retrievers were trained on eight Nvidia RTX8000  GPUs, where we set the training epoch to be 30, learning rate (lr) be 1e-5, batch size (bs) be 64, gradient accumulation step (gas) be 4.
All the readers were performed at four GTX1080 and V100 NVIDIA GPUs. For both Image-DPR and Caption-DPR,  In CReader, we set the training epoch as 3, lr as 2e-5, and batch-size as 16. In EReader, we set the training epoch as 3, lr as 1e-5, batch-size as 4, and gradient accumulation as 4.

\section{Dataset Cleaning} \label{apd:clean_corpus}

We cleaned our knowledge corpus following the steps in Section 2.2 of ~\citep{raffel2019exploring}.
Specifically, we removed the knowledge that contains any word from ``List of Dirty, Naughty, Obscene or Otherwise Bad Words''. \footnote{\url{https://github.com/LDNOOBW/List\%2Dof\%2DDirty\%2DNaughty\%2DObscene-and-Otherwise\%2DBad\%2DWords}} Knowledge that includes ``JavaScript'' and ``lorem ipsum'' is also removed. We also eliminate every knowledge with curly bracket ``\{''. Such cleaning steps remove 1\% of the knowledge from the corpus, leading the clean training corpus size to be 111,412. Similarly, 1\% of the knowledge from the full corpus is removed to make the clean full corpus size under 166,390. 

After obtaining the clean corpus, we apply our best visual retriever and visual reader to the OKVQA challenge. Specifically, first, we apply Caption-DPR fetch 100 knowledge from the clean corpus, then ask EReader to predict the answer. When using the clean training corpus, the accuracy is 39.15, and the accuracy of the clean full corpus is 44.98.

\end{document}